\documentclass[11pt,letterpaper]{article}
\usepackage{arabtex}
\usepackage{utf8}

\usepackage[utf8x]{inputenc}
\usepackage[hebrew,english]{babel}

\usepackage{arydshln}
\usepackage{textcomp}

\usepackage{emnlp2016}
\usepackage{times}
\usepackage{latexsym}

\emnlpfinalcopy

\def\emnlppaperid{190}

\usepackage[usenames,dvipsnames]{color}
\usepackage{bbding}
\usepackage{graphicx}
\usepackage{multirow}
\usepackage{booktabs}
\usepackage{setspace}

\newcommand{\COMMENT}[1] {}

\def\wit3{{\scshape WIT$^{\bf 3}$}}

\title{An Arabic-Hebrew parallel corpus of TED talks}

\author{Mauro Cettolo  \\FBK, Trento, Italy \\ cettolo@fbk.eu}

\date{}
\begin{document}

\setcode{utf8}
\maketitle

\begin{abstract}
We describe an Arabic-Hebrew parallel corpus of TED talks built upon
\wit3, the Web inventory that repurposes the original content of the
TED website in a way which is more convenient for MT researchers.

The benchmark consists of about 2,000 talks, whose subtitles in Arabic
and Hebrew have been accurately aligned and rearranged in sentences,
for a total of about 3.5M tokens per language. Talks have been
partitioned in train, development and test sets similarly in all
respects to the MT tasks of the IWSLT 2016 evaluation campaign.

In addition to describing the benchmark, we list the problems
encountered in preparing it and the novel methods designed to solve
them. Baseline MT results and some measures on sentence length are
provided as an extrinsic evaluation of the quality of the benchmark.

\end{abstract}

\section{Introduction}

TED is a nonprofit organization that ``invites the world's most
fascinating thinkers and doers [...] to give the talk of their
lives''. Its website\footnote{www.ted.com} makes the video
recordings of the best TED talks available under a Creative Commons
license. All talks have English captions, which have also been
translated into many languages by volunteers worldwide.

\wit3~\cite{cettolo:2012:EAMT}\footnote{wit3.fbk.eu} is a Web
inventory that offers access to a collection of
TED talks, redistributing the original TED website contents through
yearly releases. Each release is specifically prepared for supplying
train, development and test data to participants at MT and
SLT tracks of the evaluation campaign organized by the International
Workshop on Spoken Language Translation (IWSLT).

Despite almost all English subtitles of TED talks have been translated
into both Arabic and Hebrew, no IWSLT evaluation campaign proposed
Arabic-Hebrew as an MT task. Actually, early releases of \wit3
distributed train data for hundreds of pairs, including
Arabic-Hebrew. Nevertheless, those linguistic resources were prepared
by means of a totally automatic procedure, with only rough sanity
checks, and include talks available at that time.

Given the increasing interest in the Arabic-Hebrew task and the many
more TED talks translated into the two languages available to date, we
decided to prepare a benchmark for Arabic-Hebrew. We exploited \wit3
for collecting raw data; moreover, for making the dissemination of
results easier to users, we borrowed the partition of TED talks into
train, development and test sets adopted in the IWSLT 2016 evaluation
campaign.  

The Arabic-Hebrew benchmark is available for download at:
\begin{center}
wit3.fbk.eu/mt.php?release=2016-01-more
\end{center}

In this paper we present the benchmark, list the problems
encountered while developing it and describe the methods applied to
solve them. Baseline MT results and specific measures on the train
sets are given as an extrinsic evaluation of the quality of the
generated bitext.

\section{Related Work}

To the best of our knowledge, to date the richest collection of
publicly available Arabic-Hebrew parallel corpora is part of the OPUS
project;\footnote{opus.lingfil.uu.se} in total, it provides more than
110M tokens per language subdivided into 5 corpora, OpenSubtitles2016
being by far the largest. The OpenSubtitles2016
collection~\cite{Lison:LREC:2016}\footnote{opus.lingfil.uu.se/OpenSubtitles2016.php}
provides parallel subtitles of movies and TV programs made available
by the Open multilanguage subtitle
database.\footnote{www.opensubtitles.org} The size of this corpus
makes it outstandingly valuable; nevertheless, the translation of such
kind of subtitles is often less literal than in other domains (even
TED), likely affecting the accuracy of the fully automatic processing
implemented for parallelizing the Arabic and Hebrew subtitles.

Another Arabic-Hebrew corpus we are aware of is that manually prepared
by~\newcite{Shilon:MT:2012} for development and evaluation purposes;
no statistics on its size is provided in the paper, nor it is publicly
available; according to~\newcite{Kholy:15}, it consists
of some hundred of sentences, definitely less than those included in
our benchmark.

\section{Parallel Corpus Creation}

English subtitles of TED talks are segmented on the basis of the
recorded speech, for example in correspondence of pauses, and to fit
the caption space, which is limited; hence, in general, the single
caption does not correspond to a sentence.

The natural translation unit considered by human translators is the
caption, as defined by the original transcript.  While translators can
look at the context of the single captions, arranging this way any NLP
task -- in particular MT -- would make it particularly difficult,
especially when word re-ordering across consecutive captions
occurs. For this reason, we aim to re-build the original sentences,
thus making the NLP/MT tasks more realistic.

\subsection{Collection of talks}

For each language, \wit3 distributes  a single XML file which
includes all talks subtitled in that language; the XML format is
defined in a specific
DTD.\footnote{wit3.fbk.eu/archive/XML\_releases/wit3.dtd}
Thus, we did not need to crawl any data, as we could download the
three XML files of Arabic, Hebrew and English, available at
wit3.fbk.eu/mono.php?release=XML\_releases.

\subsection{Alignment issues}
\label{sec:AlIssue}

Even if translators volunteering for TED translated the English
captions as pointed out above, sometimes they did not adhere to the
source segmentation.  For example, in talk
n.~2357,\footnote{www.ted.com/talks/christine\_sun\_kim\_the\_enchanting\_mu{\linebreak}sic\_of\_sign\_language}
the English subtitle:

\begin{tabbing}
\hspace{2mm}\=\em French sign language was brought to America dur- \\
\> \em ing the early 1800s,
\end{tabbing} 

\noindent
is put between timestamps 53851 and 59091, while the corresponding Arabic translation is split into two subtitles:

\smallskip
      \begin{arabtext}\small لغة الإشارة الفرنسيه اعتُمِدت في امريكا\end{arabtext}
      \begin{arabtext}\small في أوائل القرن التاسع عشر\end{arabtext}
\smallskip
\smallskip

\noindent
which span the audio recording from 53851 to 56091 and from 56091
to 59091, and literally mean ``French sign language was brought
to America" and ``in the early nineteenth century'', respectively.

Even though the differences produced by translators involve a small
amount of captions (0.5\% in the Arabic-Hebrew case), these
differences affect a relevant number of talks (9\%) and in them all
subtitles following those differently segmented are desynchronized,
making the re-alignment indispensable.

\subsection{Sentence rebuilding issues}
\label{sec:RebIssue}

\begin{figure}[ht]
\centering
\begin{minipage}[h]{\columnwidth}
\begin{tabbing}
\hspace{2mm}\=\em I'd like to invite you to close your eyes.\\
\>\em Imagine yourself [...] front door of your home.\\
\>\em I'd like you to notice the color of the door,\\
\>\em the material that it's made out of.
\end{tabbing} 
\hspace{20mm}\includegraphics[width=0.65\columnwidth]{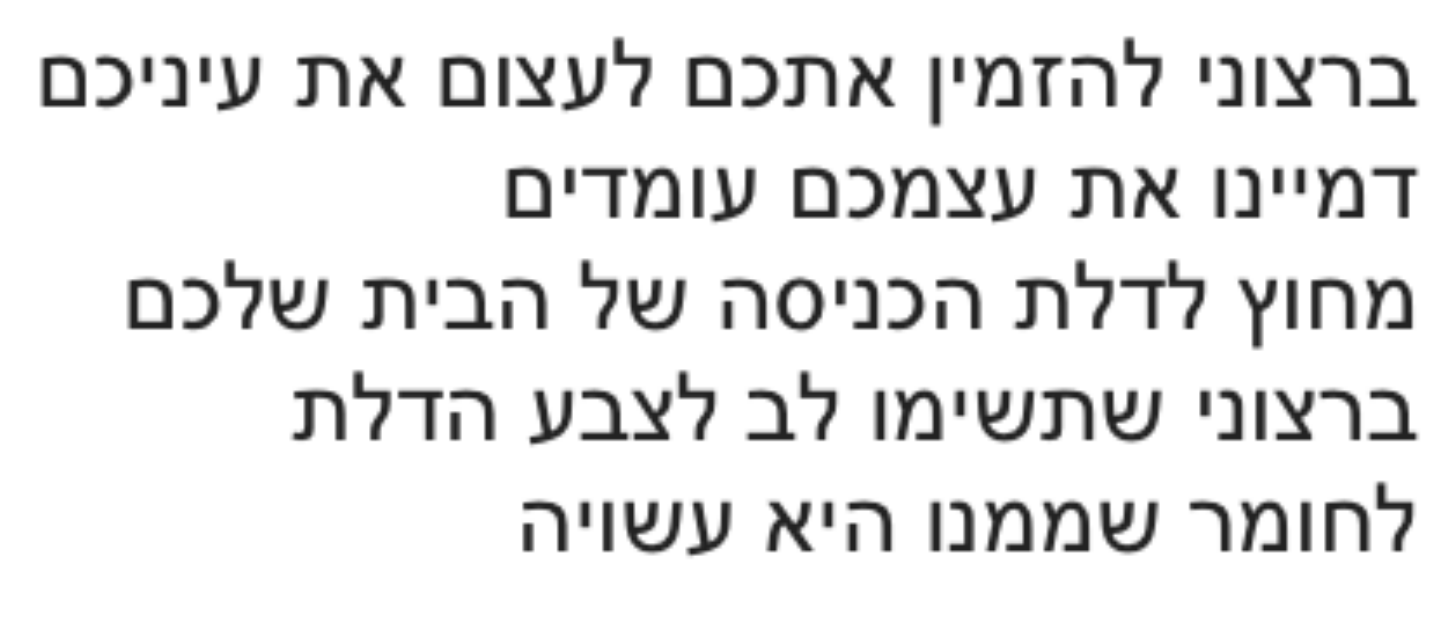}
\end{minipage}
\caption{Example of original English and Hebrew subtitles.}
\label{fig:1443}
\end{figure}

For rebuilding sentences, \wit3 automatic tools leverage strong
punctuation. Unfortunately, Arabic spelling is often inconsistent in
terms of punctuation, as both Arabic UTF8 symbols and ASCII English
punctuation symbols are used. Even worse, both in Arabic and Hebrew
translations the original English punctuation is often
ignored. An extreme case is talk
n.~1443\footnote{www.ted.com/talks/joshua\_foer\_feats\_of\_memory\_anyone\_{\linebreak}can\_do}
where 97\% of full stops at the end of English subtitles does not
appear in the Hebrew translations. The initial subtitles of that
talk are shown in Figure~\ref{fig:1443}.

Note that the strong punctuation of the English side never appears in
the Hebrew side. This example also shows the misalignment between
subtitles discussed in Section~\ref{sec:AlIssue}, being the second
English subtitle split into two Hebrew subtitles (the second and the
third).


The two issues discussed above led us to believe that trying to
directly align Arabic and Hebrew subtitles and rebuild sentences can
fail in so many cases that the overall quality of the final bitext can
be seriously affected.  We thus designed a two-stage process in which
English plays the role of the pivot. The two stages are described in
the two following sections.

\subsection{Pivot-based alignment}
\label{sec:pivotAl}

\begin{figure}[ht]
\centering
\includegraphics[width=0.9\columnwidth]{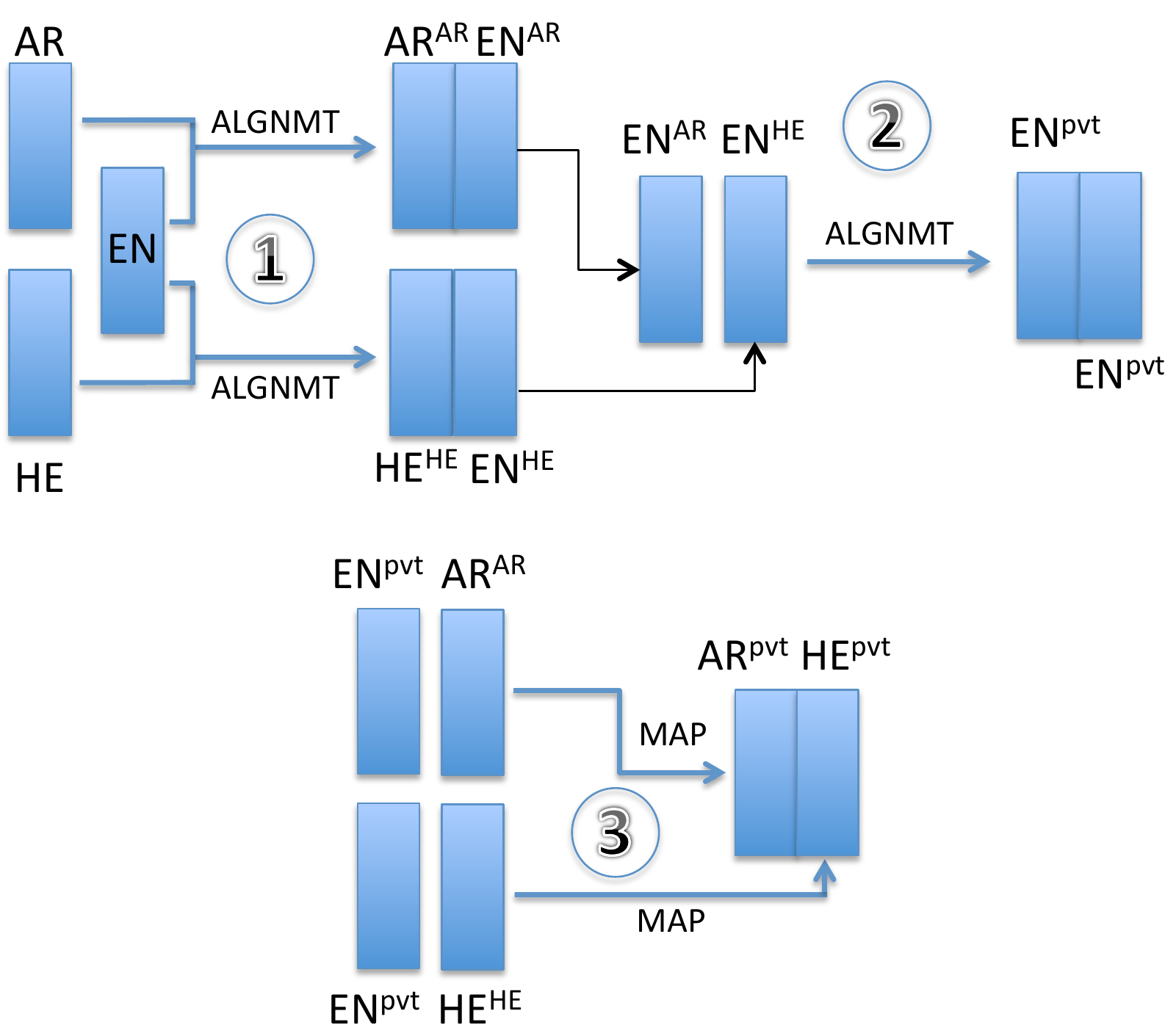}
\caption{Pivot-based alignment procedure.}
\label{fig:pivot}
\end{figure}

The alignment of Arabic and Hebrew subtitles is obtained by means of
the algorithm sketched in Figure~\ref{fig:pivot}.

The starting point are the XML files of subtitles in the three
languages.  English is aligned to Arabic and to Hebrew (step 1 in the
figure) by means of two independent runs of Gargantua, a sentence
aligner described in~\cite{braune-fraser:2010:POSTERS}. As discussed
in Section~\ref{sec:AlIssue}, the two resulting English sides can be
desynchronized, as indeed it is: in one third of the talks, the number
of subtitles differs in the two alignments. Then, Gargantua is run
again to align the two desynchronized English sides (step 2); now, the
two maps from English to English are used to rearrange the Arabic and
Hebrew sides (step 3), that at this point are aligned.

The automatic procedure drafted above is not error-proof; while
measuring failures in steps 1 and 3 of the algorithm is unfeasible
without gold references, it is simple for step 2, which should output
two perfectly equal English sides; on the contrary, about 2,000 aligned
English subtitles (out of 530,000, 0.4\%) are different, involving
less than 0.5\% of all words. Even if not immune from mistakes, the
error rate is so small that can be accepted in our context.

\subsection{Pivot-based sentence rebuilding}

The last stage in the preparation of the Arabic-Hebrew parallel corpus
is the rebuilding of sentences from the aligned subtitles. As
discussed in Section~\ref{sec:RebIssue}, we cannot rely on strong
punctuation occurring in the texts of these two languages. Once again,
the English side comes in handy. In fact, the procedure presented in
Section~\ref{sec:pivotAl} outputs the lists of Arabic, Hebrew and
English subtitles perfectly synchronized. Since punctuation marks on
the English side are reliable, sentences in the three languages are
regenerated by concatenating consecutive captions until a proper
punctuation mark is detected on the English side.

\section{Data Partitioning and Statistics}

As of April 2016, \wit3 distributes the English transcriptions of 2085
TED talks; for 2029 of them the Arabic translation is available, while
2065 have been translated into Hebrew.

The talks common to the three languages (2023) have been processed by
means of the alignment/sentence-rebuilding procedure described in the
previous section. They have been arranged in train/development/test
sets following the same partitioning adopted in MT tasks of the IWSLT
2016 evaluation campaign.  Following the IWSLT practice, the talks
that are included in evaluation sets of any past evaluation campaign
based on TED talks have been removed from the train sets, even if they
do not appear in dev/test sets of this Arabic-Hebrew release.  For
this reason, the release has a total number of aligned talks (1908)
smaller than 2023.

 \begin{table}[t]
\small
  \centering
  \begin{tabular}{|lc||r|r|}
    \hline
\multicolumn{1}{|c}{data set} & \multicolumn{1}{c||}{lang} &  \multicolumn{1}{c|}{sent} & \multicolumn{1}{c|}{\hspace*{0.2em}tokens\hspace*{0.2em}} \\
    \hline
    \multicolumn{4}{c}{} \\[-0.95em]
    \hline
     \multirow{2}{*}{train} & Ar & 153k  & 3.55M \\
    \cline{2-4}
                            & He & 223k  & 3.58M \\
    \hline
  \end{tabular}
  \caption{Monolingual resources}
  \label{tab:resource:monolingual}
 \end{table}
\begin{table}[t]
\small
  \tabcolsep5pt
  \centering
  \begin{tabular}{|l||r|rr|r|}
    \hline
\multicolumn{1}{|c|}{data} &  \multicolumn{1}{c|}{\multirow{2}{*}{sent}} & \multicolumn{2}{c|}{\hspace*{0.2em}tokens\hspace*{0.2em}} & \multicolumn{1}{c|}{\multirow{2}{*}{talks}} \\
\multicolumn{1}{|c|}{set}  &    & \multicolumn{1}{|c}{Ar} & \multicolumn{1}{c|}{He} & \\
    \hline\hline
train         &   215k   &  3.43M    & 3.38M     & 1799 \\
    \hline
dev2010       &   874    &  15,5k    & 15,0k     &    8 \\
    \hdashline
tst2010       & 1,549    &  24,6k    & 23,8k     &   11 \\
    \hdashline
tst2011       & 1,425    &  21,6k    & 21,1k     &   16 \\
    \hdashline
tst2012       & 1,703    &  23,3k    & 23,7k     &   15 \\
    \hdashline
tst2013       & 1,365    &  23,1k    & 22,9k     &   20 \\
    \hdashline
tst2014       & 1,286    &  19,8k    & 19,5k     &   15 \\
    \hline
tst2015       & 1,199    &  18,6k    & 18,9k     &   12 \\
    \hdashline
tst2016       & 1,047    &  16,5k    & 16,5k     &   12 \\
    \hline
total         & 225k     &  3.59M    & 3.54M     & 1908 \\
    \hline

  \end{tabular}
  \caption{Bilingual resources}
  \label{tab:resource:bilingual}
 \end{table}

Tables~\ref{tab:resource:monolingual} and~\ref{tab:resource:bilingual}
provide statistics on monolingual and bilingual corpora of the
Arabic-Hebrew release. Monolingual resources slightly extend the
bilingual train sets by including those talks that were not aligned
for some reason, e.g. the lack of translation in the other
language.

Figures refer to tokenized texts. The standard tokenization via the
tokenizer script released with the Europarl corpus~\cite{Koehn:05b}
was applied to English and Hebrew languages, while Arabic was
normalized and tokenized by means of the QCRI Arabic Normalizer
3.0.\footnote{alt.qcri.org/tools/arabic-normalizer/}

\section{Extrinsic Quality Assessment}

The most reliable intrinsic evaluation of the quality of the benchmark
would consist in asking human experts in the two languages to judge
the level of parallelism of a statistically significant amount of
randomly selected bitext. Since we could not afford it, we performed a
series of extrinsic checks based on both MT runs and measures on the
train sets.

\subsection{MT baseline performance}

Performance of baseline MT systems on two test sets have been
measured. The assumption behind this indirect check is that the better
the MT performance, the higher the quality of the train data (and by
extension of the whole benchmark).

\begin{table}[t]
\small
  \tabcolsep2.5pt
  \centering
  \begin{tabular}{|cc||cc|cc|}
    \hline
train   & test & \multicolumn{2}{|c|}{Ar$\rightarrow$He} & \multicolumn{2}{|c|}{Ar$\leftarrow$He} \\
rebuild.       & rebuild. & tst2012 & tst2013 & tst2012 & tst2013\\
    \hline
none         & strngP & 11.3    & 10.2    &  9.9    & 9.6  \\
strngP       & strngP & 11.3    & 10.3    &  10.4   & 9.7  \\
pivot        & strngP & 11.4    & 10.5    &  10.5   & 9.7  \\
    \hdashline
GT           & pivot  & 12.3    & 12.2    &  9.6    & 10.9 \\
\em pivot        & \em pivot  & \em 12.0    & \em 10.4    &  \em 10.6   & \em 9.8  \\
    \hline
  \end{tabular}
  \caption{BLEU scores of MT baseline systems vs. different sentence rebuilding methods. Google Translate (GT) performance is given for the sake of comparison.}
  \label{tab:mt}
 \end{table}

SMT systems were developed with the MMT
toolkit,\footnote{www.modernmt.eu} which builds engines on the Moses
decoder~\cite{Koehn:07}, IRSTLM~\cite{Federico:08} and {\tt
  fast\_align}~\cite{fastalign:13}.

The baseline MT engine (named {\tt pivot}) was estimated on the train
data of the benchmark; for comparison purposes, two additional MT
systems were trained on two Arabic-Hebrew bitexts built on the same
train TED talks of our benchmark but differently processed; in both,
subtitles were aligned directly, without pivoting through English;
then, in one case the original captions were kept as they are,
i.e. without any sentence reconstruction ({\tt none}); in the other
case, sentences were rebuilt by looking at the strong punctuation of
the Hebrew side, without using English as the pivot ({\tt
  strngP}). Note that the {\tt strngP} method is the one typically
used in \wit3 releases.

Table~\ref{tab:mt} collects the BLEU scores of our MT systems and of
Google Translate on {\tt tst2012} and {\tt tst2013} sets. The first
three rows refer to the test sets with sentences rebuilt on the Hebrew
strong punctuation; the last row regards the actual benchmark in all
respects. The score gaps are small but it has to be considered that
they are only due to the possible differences of
just a portion of subtitles (those desynchronized by the translators,
as discussed in Section~\ref{sec:AlIssue}) in a small fraction of
talks (9\%, again Section~\ref{sec:AlIssue}) used for training. Other
differences, like those shown in Section~\ref{sec:Example}, cannot
impact too much on the overall quality of the models.  Given such a
limited field of action, the gain yielded by the proposed approach is
even unexpected.

It is worth to note that the quality of our baseline systems is on a
par with Google Translate and with the state of the art phrase-based
and neural MT systems trained on our benchmark and described
in~\cite{Belinkov:2016:SeMaT}.

\subsection{Measurements on the train sets}

A set of measurements regarding the length of paired sentences has
been performed on the train set.  Table~\ref{tab:senlen} summarizes
the values of original subtitles ({\tt none}) and of sentences
generated by the {\tt strngP} and {\tt pivot} methods.  We see that
the variability of sentence length in the {\tt pivot} version equals
that of the original subtitles, which can be taken as the reference,
while the length of {\tt strngP} sentences vary much more. Moreover,
the amount of sentences longer than 100 tokens, which typically are
unmanageable/useless in standard processing, is four/five times lower
in {\tt pivot} case than in {\tt strngP}.

\begin{table}[t]
\small
  \tabcolsep2pt
  \centering
  \begin{tabular}{|c||cccc|cccc|}
    \hline
train   & \multicolumn{4}{|c|}{Ar} & \multicolumn{4}{|c|}{He} \\
rebuild.     & $\mu$ & $\sigma$ & max & \textperthousand$^{^{>100}}$ & $\mu$ & $\sigma$  & max & \textperthousand$^{^{>100}}$\\
    \hline
none         & 13.3 & 11.6 & 110  & 0.03 & 12.9 & 11.2 & 111  & 0.04 \\
strngP       & 19.3 & 22.6 & 2561 & 3.4  & 18.8 & 20.8 & 2294 & 3.0 \\
pivot        & 16.0 & 11.7 & 495  & 0.7  & 15.7 & 11.7 & 703  & 0.7 \\
    \hline
  \end{tabular}
  \caption{Statistics on length of train sentences for different rebuilding methods. \textperthousand$^{^{>100}}$ stands for the per thousand rate of sentences longer than 100 tokens.}
  \label{tab:senlen}
 \end{table}

\begin{table}[t]
\small
  \centering
  \begin{tabular}{|c||cc|}
    \hline
train   & \multirow{2}{*}{$\mu$} & \multirow{2}{*}{$\sigma$} \\
rebuild.     &       &          \\
    \hline
none         & 0.39 & 3.4 \\
strngP       & 0.57 & 5.1 \\
pivot        & 0.26 & 4.1 \\
    \hline
  \end{tabular}
  \caption{Statistics on the length difference between the Arabic and Hebrew train sentences for different rebuilding methods.}
  \label{tab:sendifflen}
 \end{table}

Finally, Table~\ref{tab:sendifflen} provides the mean and the standard
deviation of the difference of the number of tokens between Arabic and Hebrew
subtitles. Also here the statistics on original subtitles ({\tt none})
can be assumed to be the gold reference, and again the {\tt pivot}
version is preferable to the {\tt strngP} version.

\subsection{Example}
\label{sec:Example}

Here we show how the  three methods {\tt none}, {\tt strngP} and
{\tt pivot} process the example of Figure~\ref{fig:1443}.  For the
sake of readability, only the English translation is given.

All methods properly align original captions.  Differences come
from the sentence rebuilding. 

By definition, {\tt none} keeps the five original Ar/He subtitles:

\begin{tabbing}
\hspace{2mm}\=\em I'd like to invite you to close your eyes.\\
\>\em Imagine yourself standing\\
\>\em outside the front door of your home.\\
\>\em I'd like you to notice the color of the door,\\
\>\em the material that it's made out of.
\end{tabbing} 

\noindent {\tt strngP}, misled by the absence of strong
punctuation on the Hebrew side, appends together the five subtitles
(and many more) into one long ``sentence'':

\begin{tabbing}
\hspace{2mm}\=\em I'd like to invite [...] that it's made out of. [...]
\end{tabbing} 

\noindent {\tt pivot} is instead able to properly reconstruct
sentences from the original captions:

\begin{tabbing}
\hspace{2mm}\=\em I'd like to invite you to close your eyes.\\
\>\em Imagine yourself standing [...] door of your home.\\
\>\em I'd like you to notice [...] that it's made out of.
\end{tabbing} 

\noindent so providing the best segmentation from a linguistic point
of view.

\section{Summary}

In this paper we have described an Arabic-Hebrew benchmark built on
data made available by \wit3.  The Arabic and Hebrew subtitles of
around 2,000 TED talks have been accurately rearranged in sentences
and aligned by means of a novel and effective procedure which relies
on English as the pivot. The talks count a total of 225k
sentences and 3.5M tokens per language and have been partitioned in
train, development and test sets following the split of the MT tasks
of the IWSLT 2016 evaluation campaign.

\section*{Acknowledgments}

This work was partially supported by the CRACKER project, which
received funding from the European Union's Horizon 2020 research and
innovation programme under grant no. 645357.

The author wants to thank Yonatan Belinkov for providing invaluable
suggestions in the preparation of the benchmark.

\bibliographystyle{emnlp2016}

\begin{thebibliography}{}

\bibitem[\protect\citename{Belinkov and Glass}2016]{Belinkov:2016:SeMaT}
Yonatan Belinkov and James Glass.
\newblock 2016.
\newblock Large-scale Machine Translation between Arabic and Hebrew: Available
  Corpora and Initial Results.
\newblock In {\em Proc. of SeMaT}, Austin, US-TX.

\bibitem[\protect\citename{Braune and Fraser}2010]{braune-fraser:2010:POSTERS}
Fabienne Braune and Alexander Fraser.
\newblock 2010.
\newblock Improved Unsupervised Sentence Alignment for Symmetrical and
  Asymmetrical Parallel Corpora.
\newblock In {\em Proc. of Coling 2010: Posters}, pp. 81--89, Beijing, China.

\bibitem[\protect\citename{Cettolo \bgroup et al.\egroup
  }2012]{cettolo:2012:EAMT}
Mauro Cettolo, Christian Girardi, and Marcello Federico.
\newblock 2012.
\newblock {WIT$^3$: Web Inventory of Transcribed and Translated Talks}.
\newblock In {\em Proc. of EAMT}, pp. 261--268, Trento, Italy.

\bibitem[\protect\citename{Dyer \bgroup et al.\egroup }2013]{fastalign:13}
Chris Dyer, Victor Chahuneau, and Noah~A. Smith.
\newblock 2013.
\newblock A Simple, Fast, and Effective Reparameterization of IBM Model 2.
\newblock In {\em Proc. of NAACL}, pp. 644--648, Atlanta, US-GA.


\bibitem[\protect\citename{El~Kholy and Habash}2015]{Kholy:15}
Ahmed El~Kholy and Nizar Habash.
\newblock 2015.
\newblock Morphological Constraints for Phrase Pivot Statistical Machine
  Translation.
\newblock In {\em Proc. of MT Summit XV, vol.1: MT Researchers' Track},
  pp. 104--116, Miami, US-FL.

\bibitem[\protect\citename{Federico \bgroup et al.\egroup }2008]{Federico:08}
Marcello Federico, Nicola Bertoldi, and Mauro Cettolo.
\newblock 2008.
\newblock {IRSTLM: an Open Source Toolkit for Handling Large Scale Language
  Models}.
\newblock In {\em Proc. of Interspeech}, pp. 1618--1621, Brisbane,
  Australia.

\bibitem[\protect\citename{Koehn \bgroup et al.\egroup }2007]{Koehn:07}
Philipp Koehn, Hieu Hoang, Alexandra Birch, Chris Callison-Burch, Marcello
  Federico, Nicola Bertoldi, Brooke Cowan, Wade Shen, Christine Moran, Richard
  Zens, Chris Dyer, Ondrej Bojar, Alexandra Constantin, and Evan Herbst.
\newblock 2007.
\newblock {Moses: Open Source Toolkit for Statistical Machine Translation}.
\newblock In {\em Proc. of ACL: Demo and Poster
  Sessions}, pp. 177--180, Prague, Czech Republic.

\bibitem[\protect\citename{Koehn}2005]{Koehn:05b}
Philipp Koehn.
\newblock 2005.
\newblock Europarl: A Parallel Corpus for Statistical Machine Translation.
\newblock In {\em Proc. of MT Summit X}, pp. 79--86, Phuket, Thailand.

\bibitem[\protect\citename{Lison and Tiedemann}2016]{Lison:LREC:2016}
Pierre Lison and Jörg Tiedemann.
\newblock 2016.
\newblock Opensubtitles2016: Extracting Large Parallel Corpora from Movie and
  TV Subtitles.
\newblock In {\em Proc. of LREC}, pp. 923--929, Portoro\v{z}, Slovenia.

\bibitem[\protect\citename{Shilon \bgroup et al.\egroup }2012]{Shilon:MT:2012}
Reshef Shilon, Nizar Habash, Alon Lavie, and Shuly Wintner.
\newblock 2012.
\newblock Machine Translation between Hebrew and Arabic.
\newblock {\em Machine Translation}, 26(1-2):177--195, March.

\end{thebibliography}

\end{document}